\documentclass[conference]{IEEEtran}
\IEEEoverridecommandlockouts

\usepackage{cite}
\usepackage{amsmath,amssymb,amsfonts,amsthm,mathtools}
\usepackage[alphabetic,backrefs,lite,nobysame]{amsrefs}
\usepackage{algorithm2e}
\usepackage{tikz-cd}
\usepackage{graphicx}
\usepackage{listings}
\usepackage{textcomp}
\usepackage{xcolor}
\usepackage{subcaption}

\newcommand\pder[2][]{\ensuremath{\frac{\partial#1}{\partial#2}}} 
\newcommand{\R}{\mathbb{R}}
\newcommand\blfootnote[1]{%
	\begingroup
	\renewcommand\thefootnote{}\footnote{#1}%
	\addtocounter{footnote}{-1}%
	\endgroup
}

\DeclareMathOperator{\MSE}{MSE}
\DeclareMathOperator{\PD}{PD}
\DeclareMathOperator{\CONV}{conv}

\def\BibTeX{{\rm B\kern-.05em{\sc i\kern-.025em b}\kern-.08em
		T\kern-.1667em\lower.7ex\hbox{E}\kern-.125emX}}

\theoremstyle{definition}

\numberwithin{figure}{section}

\graphicspath{{./img/}}

\begin{document}

	\title{A Fast and Robust Method for Global Topological Functional Optimization
		\thanks{The first and third authors were partially supported by the Air Force Office of Scientific Research under the  grant ``Geometry and Topology for Data Analysis and Fusion",  AFOSR FA9550-18-1-0266. The second author was partially supported by the National Science Foundation under the grant ``HDR TRIPODS: Innovations in Data Science: Integrating Stochastic Modeling, Data Representations, and Algorithms", NSF CCF-1934964. We are grateful to Jay Hineman for technical discussions.}
	}
	
	\author{
		Elchanan Solomon$^*$\\
		Department of Mathematics, \\
		Duke University\\
		Durham, USA \\
		yitzchak.solomon@duke.edu
		\and
		Alexander Wagner$^*$\\
		Department of Mathematics, \\
		Duke University\\
		Durham, USA \\
		alexander.wagner@duke.edu
		\and
		Paul Bendich\\
		Department of Mathematics, Duke University\\
		Geometric Data Analytics\\
		Durham, USA \\
		paul.bendich@duke.edu
	}
	
	\maketitle
	
	\begin{abstract}
		Topological statistics, in the form of persistence diagrams, are a class of shape descriptors that capture global structural information in data. The mapping from data structures to persistence diagrams is almost everywhere differentiable, allowing for topological gradients to be backpropagated to ordinary gradients. However, as a method for optimizing a topological functional, this backpropagation method is expensive, unstable, and produces very fragile optima. Our contribution is to introduce a novel backpropagation scheme that is significantly faster, more stable, and produces more robust optima. Moreover, this scheme can also be used to produce a stable visualization of dots in a persistence diagram as a distribution over critical, and near-critical, simplices in the data structure.
	\end{abstract}
	
	\blfootnote{$^*$ Equal contribution}

In its early days, topological data analysis (TDA) was viewed as being in competition with other methods and models in data science, and much of TDA research proceeded independently from the state-of-the-art in the machine learning space. In recent years, however, the role of TDA as a component in a larger data analysis pipeline has come to the forefront. One can divide the literature into the following streams:

\begin{enumerate}
	\item Using TDA to extract features from data that are then fed into standard machine learning or statistics pipelines. Cf. \cite{bendich2016persistent}, \cite{brown2009nonlinear}, and \cite{gamble2010exploring}.
	\item Using topological signatures as measures of model complexity. Cf. \cite{gebhart2019characterizing}, \cite{guss2018characterizing}, \cite{corneanu2019does}, and \cite{rieck2018neural}.
	\item Designing neural network architectures that can handle topological signatures. Cf. the PersLay architecture of \cite{carriere2019general}.
	\item Incorporating topological terms into classical loss functions. Cf. \cite{chen2018topological} and \cite{hu2019topology}.
\end{enumerate}

As this work belongs to the final stream above, let us consider the prior work in greater depth. \cite{chen2018topological} propose adding a regularizer term to the loss function of a complex model that penalizes the topological complexity of the decision boundary. They introduce an efficient algorithm for computing the gradients of this topological penalty, implement it in conjunction with a standard kernel classifier, and demonstrate improved results for both synthetic and real-world data sets. The computational tractability of their approach comes from the fact that the homological dimension of interest is zero, where persistence computations are particularly fast.

\cite{hu2019topology} propose a novel framework for building a neural network that maps an image to its segmentation. They introduce a topological loss into the training phase by asking that the model output approximate the ground truth segmentation in a metric that combines cross-entropy and the $2$-Wasserstein distance on persistence diagrams. To alleviate the instability of topological backpropagation and the relatively expensive computational cost of persistent homology, their framework works with one single, small patch of the image at a time. Experiments on natural and biomedical image datasets demonstrate that the incorporation of topology provides quantitatively superior results across a host of measures.

The focus of this paper is not classification or segmentation, but topological functional optimization. That is, our goal is to optimize a functional on the space of shapes that has both a classical, machine-learning component (approximating a fixed, input image in mean squared error (MSE), cross entropy, etc.) and a topological component, e.g. $\alpha \Phi(\PD(f)) + (1-\alpha) \MSE(f,f_0)$. Our proposed framework is naturally unsupervised and can accept a wide variety of user-specified functionals.  Moreover, it applies to functions defined on arbitrary simplicial complexes.

Here is a sample list of useful image optimization tasks covered by our framework:
\begin{itemize}
	\item {\bf Topologically accurate signal downsampling}. It is often prohibitively expensive to transmit large signals. One can cast downsampling as the problem of mapping a signal into a lower-dimensional space (either a shorter signal or a signal belonging to a simple parametrized family) while minimizing some measure of distortion. By incorporating a topological loss into this optimization task, we can ensure that our downsampling preserves key structural features that may be important for further analysis and classification. Cf. \cite{poulenard2018topological}. 
	
	A possible concern is that downsampling may affect the scale of the topological structures picked up. However, for many purposes, local topology is indicative of noise and presents an obstruction to the learning task. For example, in the cell segmentation task shown in Figure \ref{fig:cellsegmentation}, the segmentation is accomplished by optimizing for $1$-dimensional homology on large scales. The cell images contain small-scale $1$-dimensional homology that correspond either to noise or to local structures that are not entire cells. Moreover, there already exist effective techniques for optimizing local topology \cite{hu2019topology}.
	
	\item {\bf Image simplification}. Topological data can be used as a measure of image complexity. By defining a functional that penalizes small-scale topological features, we obtain a scheme for removing topological noise from images. Cf. \cite{10.1145/1137856.1137878}.
	
	\item {\bf Enforcing correct topology}. In contrast with the prior example, there are settings in which we believe an image ought to exhibit particular topology at a given scale, such as a certain number of connected components, the existence of a cycle or void, etc. By defining a functional that penalizes distance from this prescribed topology, we can produce a modified image with the correct topology.
\end{itemize}  

The challenges of topological functional optimization are three-fold: (1) topological gradients are very expensive to compute, (2) the mapping from topological gradients to ordinary gradients (defined on the image space) is extremely unstable, and (3) topology can be made or broken by changing individual pixels, so the optima produced via straightforward gradient descent are fragile. To address these difficulties, we introduce a novel topological backpropagation scheme that is faster, more stable, and produces more robust optima than traditional methods.

Section \ref{sec:topbackprop} reviews the literature on topological backpropagation and demonstrates how gradients in the space of persistence diagrams can be pulled back to produce gradients on the original data structure. This is followed by an analysis that explains why the traditional method is unstable, slow to compute, and produces undesirable optima. In Section \ref{sec:smearing}, we introduce our novel approach to topological backpropagation via \emph{smearing} the topological loss. This smearing procedure requires computing the topological statistics of many approximates of our data. We then introduce \textbf{STUMP}, our scheme for quickly generating these approximates, combining their gradients, and further stabilizing the result. In Section \ref{sec:experiments}, we consider three synthetic optimization tasks and compare the results of ``vanilla" topological optimization with our new, smeared approach. We perform a robustness and speed analysis, demonstrating that our method outperforms the traditional one in both metrics. In Section \ref{sec:extensions}, we discuss how our pipeline can be used to provide stable visualizations of dots in a persistence diagram and a simple generalization of our pipeline to point cloud data.

\section{Topological Backpropagation}	
\label{sec:topbackprop}

\subsection{Persistent Homology}
The content of this paper assumes familiarity with the concepts and tools of persistent homology. Interested readers can consult the articles of \cite{carlsson2009topology} and \cite{ghrist2008barcodes} and the textbook of \cite{edelsbrunner2010computational}. However, we provide a brief recapitulation here. 

Given a simplicial complex $D$, a filtration is a function $f: D \to \R$ such that $f^{-1}((-\infty, t])$ is a simplicial complex for every $t$. The persistence diagram of $f$, denoted $\PD(f)$, is then a multiset in $\{(b, d) \in \R^2\ |\ b < d \}$ that records the birth and death times of homological features as the parameter $t$ varies. The $p$-Wasserstein metrics, or Bottleneck metric when $p = \infty$, are popular metrics on the space of persistence diagrams. These metrics are similar to the usual Wasserstein metrics with the important caveat that mass can be freely added to or removed from the diagonal. When we consider loss functionals $\Phi$ defined on persistence diagrams $\PD(f)$, $\Phi$ will typically be the $p$-Wasserstein distance from a thresholded $\PD(f)$ to the empty diagram raised to the $p$ power or negative this value. Explicitly, if $\PD(f) = \{(b_i, d_i)\}_{i=1}^n$ then $\Phi(\PD(f)) = \sum_{i \in I} |b_i - d_i|^p$ where $I = \{i\ |\ d_i - b_i > \epsilon \}$ for some threshold value $\epsilon$. We will refer to these functionals $\Phi$ as the $p$-Wasserstein norms in the sequel.

The incorporation of persistent homology into model training is based on three properties of persistent homology:
\begin{itemize}
	\item (semantic) Topological data can be used to measure a host of important, yet abstract, concepts in data analysis: noise, connectivity, consistency, shape, boundary, scale, dimension, etc.
	\item (computational) Persistent homology can be computed efficiently.
	\item (submersion) The differential of the map from model parameters to persistence diagrams has full rank almost everywhere. In the language of differential topology, such a map is called a submersion.
\end{itemize}

It is this final, submersion property that allows for the backpropagation of topological gradients to gradients in the parameter space of the model. We now make this backpropagation scheme precise. Let our model $g$ be parameterized by a set of parameters $\alpha_j$, let $\{b_i,d_i\}_{i \in I}$ be the set of birth-death times of the persistence diagram $\PD(g)$, and let $\Phi$ be a functional on the space of persistence diagrams. In order to optimize $\Phi$ as a function of the model parameters $\alpha_j$, we need to be able to compute the partial derivatives:
\[\pder[b_i]{\alpha_j} \,\,\,\,\,\,\,\,\, \mbox{and} \,\,\,\,\,\,\,\,\, \pder[d_i]{\alpha_j}.  \]
To compute these derivatives, one can take advantage of the pairing between birth and death times in $\PD(g)$ and critical simplices of $g$ (this pairing is well-defined in the generic setting that all critical simplices have distinct function values). When using a lower-star filtration, one can further simplify this pairing by choosing, for each critical simplex, the vertex whose additional to the filtration implied the addition of the critical simplex (there is a unique such vertex under the generic assumption that all vertex values are distinct). Let us therefore write $\pi$ to identify this mapping from birth or death times to vertices of our discretized domain $D$. Note, crucially, that $\pi$ is locally constant with respect to perturbations of the function $g$. With such a pairing, we can write the partial derivatives above in a more tractable form:
\[\pder[b_i]{\alpha_j} = \pder[g(\pi(b_i))]{\alpha_j} = \pder[g]{\alpha_j}(\pi(b_i))\] 
\[\pder[d_i]{\alpha_j} = \pder[g(\pi(d_i))]{\alpha_j} = \pder[g]{\alpha_j}(\pi(d_i))\] 

We can thus use the chain rule to deduce:
\[\pder[\Phi]{\alpha_j} = \sum_{i \in I}\pder[\Phi]{b_i}\pder[b_i]{\alpha_j} + \pder[\Phi]{d_i}\pder[d_i]{\alpha_j}  \]
\[=\sum_{i \in I}\pder[\Phi]{b_i}\pder[g]{\alpha_j}(\pi(b_i)) + \pder[\Phi]{d_i}\pder[g]{\alpha_j}(\pi(d_i)) \]

The partial derivatives $\pder[\Phi]{b_i}$ and $\pder[\Phi]{d_i}$ must be computed explicitly for the functional $\Phi$ of interest. In \cite{poulenard2018topological}, closed-forms of these derivatives are given for the special cases when $\Phi$ is the bottleneck or $2$-Wasserstein distance to a target persistence diagram.

Lastly, it is worth noting that topological backpropagation gives the user the freedom to treat birth and death simplices separately. This is often desired in practice, as will be seen in the experiments in Section \ref{sec:experiments}.

\subsection{Instability of Topological Backpropagation}
The method of topological backpropagation via persistence dot-critical vertex pairings has a number of limitations. The most crucial is that of instability. It is a well-known result in applied topology that persistence diagrams themselves are stable to perturbations of the underlying function, cf. \cite{cohen2007stability} and \cite{chazal2009proximity}. However, no such stability applies to the location of critical vertices. This failure of stability was investigated by \cite{bendich2019stabilizing} and presents itself as a challenge to many topological inverse problems. In our setting, when implementing topological backpropagation in functional optimization, this translates into unstable gradients. 

\subsection{Computational Cost of Topological Backpropagation}

Software like GUDHI \cite{gudhi:CubicalComplex} provides for fast calculation of persistence dot-critical vertex pairings. In principle, computing this pairing is no more expensive than computing persistence. To see why, let $f:D \to \mathbb{R}$ be our function, and let us assume that we are in the generic setting that $f$ is injective on the vertices of $D$. Define $F:D \to \mathbb{N}$ to map every vertex to a natural number representing the ordinal position in which it appears in the filtration induced by $f$. Thus, the vertex with the lowest $f$-value is mapped to zero, the vertex with the second $f$-value is mapped to one, and so forth. The birth and death times of the dots in the resulting persistence diagram $\PD(F)$ give the indices of their critical vertices. To transform $\PD(F)$ into $\PD(f)$, replace the birth and death indices with the $f$-values of the corresponding vertices.

\cite{morozov2005persistence} showed that the worst-case complexity of computing persistence is cubic in the number of simplices. When our simplicial complex $D$ is a triangulation of a $k$-dimensional manifold, the number of simplices grows exponentially in $k$ with the resolution of the triangulation. Thus, even for $k=2$ and $k=3$, the computation of persistence homology and the persistence dot-critical vertex pairings scales poorly in the resolution of the image. From a computational perspective, it is therefore ideal to compute as few high-resolution persistence diagrams as possible.

\subsection{Robustness of Optima}

When the output of topological optimization is perturbed, either unintentionally (in lossy communications) or intentionally (as in an adversarial attack), the topology of the image changes. An optima is robust if the value of the topological functional can only be increased by adding a substantial amount of noise, as measured in MSE. Because topological backpropagation pins the responsibility for a given topological feature on a single pixel, it tends to introduce or destroy topology in a very fragile way, as will become clear in the Experimental Results section.

\section{Smearing Topological Optimization}
\label{sec:smearing}

Consider a topological optimization task where the loss is of the form $L(f) = \alpha \Phi(\PD(f)) + (1-\alpha) \MSE(f,f_0)$. We can make this loss function more robust by associating to every function $f$ a set of approximate functions $S(f)$, equipped with a measure $\mu$, and replacing the loss with:
\[
L_{S}(f) = \alpha \int_{S(f)}\Phi(\PD(g)) d\mu(g)+ (1-\alpha) \MSE(f,f_0).
\]
Thus, the topological term of our loss function measures the ``average" topology of a set of approximates to $f$, weighted via $\mu$. Informally, we call this \emph{smearing} the topological loss over the set of approximates.

\subsection{Generating Approximates}
\label{sec:generatingapproximates}

\begin{figure}[htb!]
	\centering
	\includegraphics[scale=0.4]{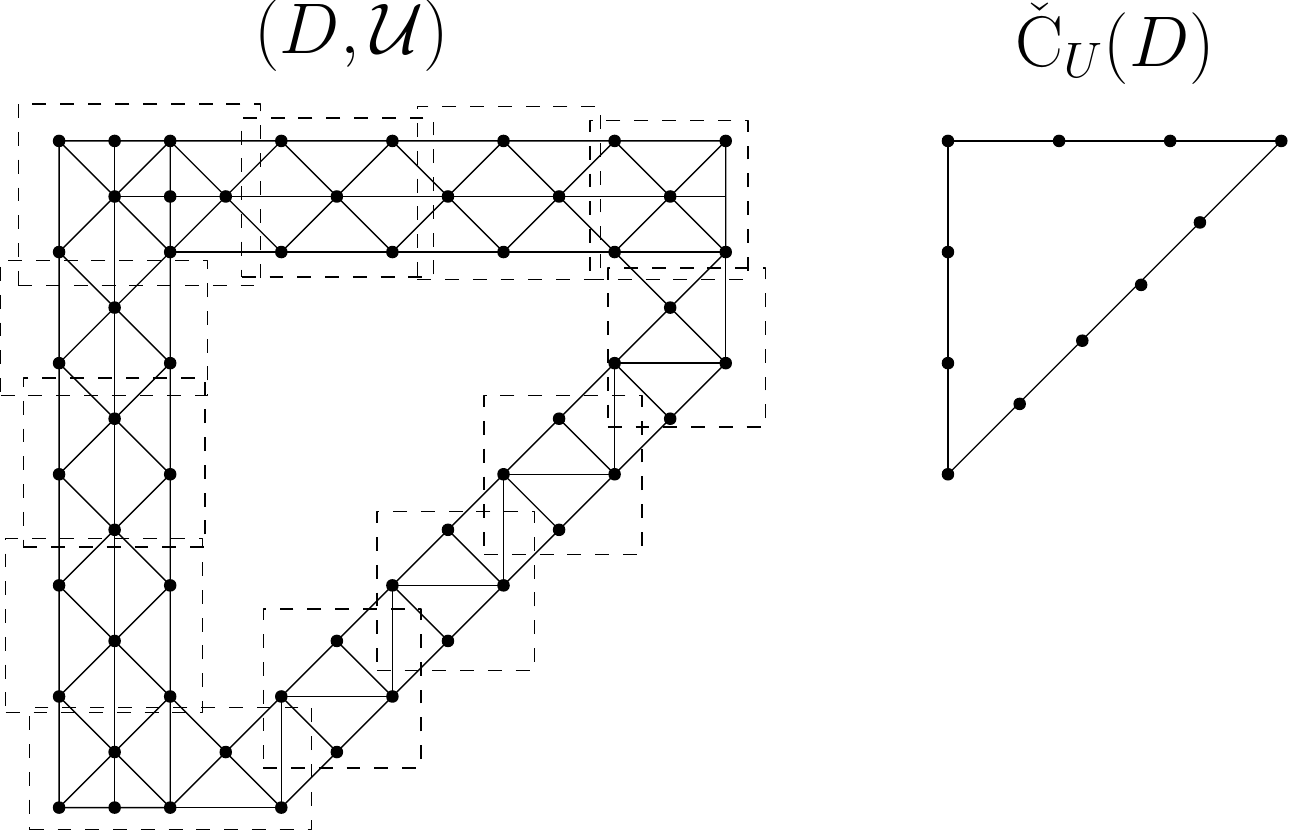}
	\caption{On the left, we see a simplicial complex $D$ with an open cover $\mathcal{U}$. The corresponding \v{C}ech complex is shown on the right. When the open sets of $\mathcal{U}$, and all their possible intersections, are contractible, the topological type of $D$ is the same as that of the \v{C}ech complex; this is the well-known Nerve Theorem.}
	\label{fig:cech}
\end{figure}

\begin{figure}[htb!]
	\centering
	\includegraphics[scale=1]{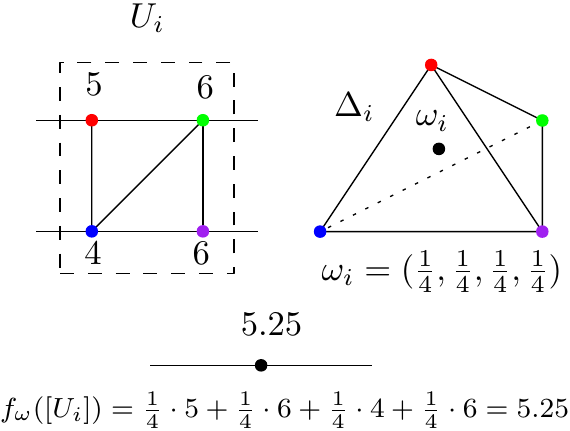}
	\caption{This figure demonstrates how the values of the approximate function $f_{\omega}$ are obtained. For a given open set $U_i$, a $\mu_i$-randomly chosen weighting $\omega_i \in \Delta_i$ prescribes a linear combination of values in $U_i$.}
	\label{fig:sample}
\end{figure}

We now introduce a general scheme for constructing sets of approximates, given a function $f: D \to \mathbb{R}$ defined on a simplicial complex\footnote{The experiments in Section \ref{sec:experiments} are actually computed using cubical complexes but are equivalent to the formalism here via the Freudenthal triangulation.}. The first ingredient is an open cover $\mathcal{U}$ of $D$. We downsample $D$ by considering the \v{C}ech complex $\mbox{\v{C}}_{\mathcal{U}}(D)$, see Figure \ref{fig:cech}. Each vertex of $\mbox{\v{C}}_{\mathcal{U}}(D)$ corresponds to an open set $U_i \in \mathcal{U}$. In order to produce a function on the \v{C}ech complex, we need a rule for averaging the set of values $\{f(v) \mid v \in U_{i}^{0}\}$ for each open set $U_i$. To that end, we associate to each open set $U_i$ the probability simplex $\Delta_i$ on its set of vertices, $U_{i}^{0}$. That is, an element $\omega_i \in \Delta_i$ is an assignment of nonnegative weights to the vertices in $U_{i}^{0}$ such that $\sum_{v \in U_{i}^{0}}\omega_{i}(v) = 1$. We write $\omega = \{\omega_i\}$ to denote a choice of element in $\Delta_i$ for each $i$, which thus gives rise to a function $f_{\omega}$ on $\mbox{\v{C}}_{\mathcal{U}}(D)$:
\[f_{\omega}([U_{i}]) = \sum_{v \in V(U_i)}\omega_{i}(v)f(v). \]
See Figure \ref{fig:sample}. The value of $f_\omega$ on a non-vertex simplex $\sigma \in \mbox{\v{C}}_{\mathcal{U}}(D)$ is defined to be the maximum value of $f_\omega$ on the vertices of $\sigma$. To specify how the elements $\omega_i \in \Delta_i$ are chosen, we pick a set of measures $\mu = \{\mu_i\}$, one for each $\Delta_i$:
\begin{enumerate}
	\item If $\mu_{i}$ is an atomic measure, concentrated on the center of the simplex $\Delta_{i}$, the resulting downsample associates to each open set $U_i$ the average of the values of $f$ on its vertices.
	\item If $\mu_{i}$ is a uniform measure on the zero-skeleton $\Delta_{i}^{0}$, a downsample $\omega$ is obtained by randomly picking a vertex $v \in U_i^{0}$ and setting $f_{\omega}([U_i]) = f(v)$.
	\item If $\mu_i$ is a uniform measure on the entirety of $\Delta_{i}$, a downsample corresponds to taking a random, normalized linear combination of the values of $f$ on the open set $U_i$.
\end{enumerate} 

The set of approximates $S(f)$ is the set of all downsampled functions $f_{\omega}$ on the \v{C}ech complex, with the measure as chosen. Since the \v{C}ech complex is smaller than the original complex, the computation of individual persistence diagrams is accelerated. 

\subsection{The Smeared Gradient}
For a fixed weighting $\omega$, the map $f \to f_{\omega}$ is linear. The chain rule then implies that:
\[\frac{d \Phi(\PD(f_\omega)) }{df} =   \frac{d \Phi(\PD(f_\omega)) }{df_\omega}\circ \omega.\]
Under mild technical assumptions that allow us to move the gradient under the integral sign, we therefore have for $\alpha = 1$:
\[\frac{dL_{S}(f)}{df} = \int_{\omega} \left( \frac{d \Phi(\PD(f_\omega)) }{df_\omega}\circ \omega \right) d\mu(\omega).  \]

This provides a simple formula for computing the gradient with respect to $f$ in terms of the gradients of the downsamples $f_{\omega}$ but is not exactly computable in practice, due to the high-dimensionality of the set $S(f)$ over which we must integrate. To approximate this integral, we can, at each step of the optimization, sample finitely many $f_{\omega}$ and compute an empirical average. An even faster approach, which we implement in practice, is to mirror stochastic gradient descent by considering a single downsampled $f_{\omega}$ at each descent step and mixing the gradients via momentum using Adam \cite{kingma2014adam}. Taken altogether, we call our pipeline \textbf{STUMP}: \textbf{S}tochastic \textbf{T}opological \textbf{U}pdates via \textbf{M}omentum and \textbf{P}ooling.

\subsection{Clarke Subdifferentials}

\begin{figure}[htb!]
	\begin{center}
		\includegraphics[width=0.4\textwidth]{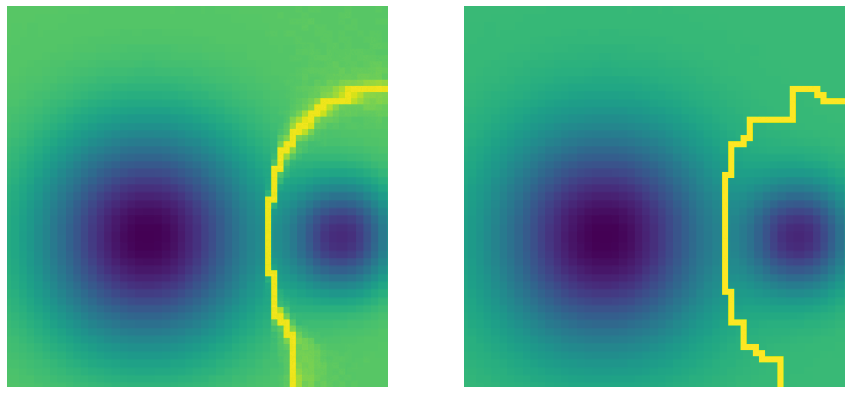}
	\end{center}
	\caption{The effect of explicitly adding noise in generating approximates. Both images are the result of near-identical optimization procedures which only differed in whether or not inputs were perturbed before each optimization step. The image on the left, for which the input had noise added, is smoother than the image on the left for which the input was not perturbed.}
	\label{fig:noisecomparison}
\end{figure}

The robustness of the optimization scheme can be further improved by considering perturbations of the initial function $f$. Here we give two heuristic motivations for adding explicit noise. The first is the qualitative effect on results of optimization. For instance, the picture on the left of Figure \ref{fig:noisecomparison} is the result of an optimization procedure that added explicit noise to the input before stochastic downsampling while the picture on the right is the result in the absence of explicit noise. 

The second argument involves the gradient sampling methodology \cite{doi:10.1137/030601296}. When minimizing an unstable function, a more robust search direction can be obtained by considering the minimum norm element of the convex hull of gradients of nearby points. More precisely, Lemma $2.1$ of \cite{doi:10.1137/030601296} states that if $G$ is a compact convex subset of $\R^d$ and $g^* \in G$ is a minimum norm element of $G$, then $d^* = -g^*/ \|g^* \|$ solves $\inf_{\| d \| \leq 1} \sup_{g \in G} \langle g, d \rangle$. In other words, $d^*$ is a minimax update direction.

The space of perturbations that will be used in the experiments in Section \ref{sec:experiments} is the cube $[-\epsilon, \epsilon]^d$, so $G$ will be the convex hull of gradients of points in $x + [-\epsilon, \epsilon]^d$. As a proxy for finding the minimum norm element of $G$, one can sample points $x_1, \dots, x_m \in x + [-\epsilon, \epsilon]^d$, compute gradients $g_i = \nabla \Phi(\PD(x_{i, \omega_i}))$ at each of these points, and find the minimum norm element of $\CONV(g_1, \dots, g_m)$, i.e.
\[
\min \| \sum_{i=1}^m c_i g_i \|^2 \text{ subject to } c \in \Delta^{m-1}.
\]
If the $g_i$ are pairwise orthogonal, the problem above has the simple solution $c_i := \| g_i \|^{-2} / (\sum_{i=1}^m \| g_i \|^{-2})$. If in addition the norm of each $g_i$ is equal, then each $c_i$ would equal $1/m$. In other words, under these two extreme assumptions, we may approximate a robust update direction by simply averaging nearby gradients. We tested the validity of these assumptions for a particular example, the starting point of the smear optimization for the blobs experiment in Section \ref{sec:experiments}.

\begin{figure}[htb!]
	\begin{center}
		\includegraphics[width=0.5\textwidth]{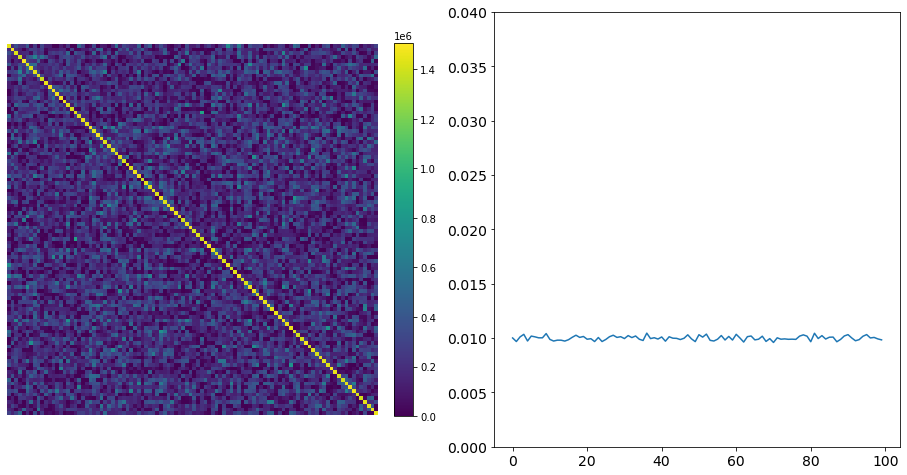}
	\end{center}
	\caption{The $ij$-entry of the matrix on the left is $\langle g_i, g_j \rangle$. The graph on the right shows the values of $c_1, \dots, c_{100}$.}
	\label{fig:clarke}
\end{figure}

Figure \ref{fig:clarke} shows the Gram matrix on the left and the values of the $c_i$'s defined above on the right when $m = 100$. Note that the Gram matrix is somewhat diagonal and the values of the $c_i$ fluctuate very tightly around $0.01 = 1/m$. The degree of orthogonality among the $g_i$'s corresponds to the degree of instability of persistence dot-critical vertex pairings. On the other hand, the stability of the $c_i$'s reflects the stability of persistence diagrams to perturbation.

\section{Experimental Results\protect\footnote{https://github.com/aywagner/TDA-smear}}
\label{sec:experiments}
We now consider a number of topological
optimization tasks and compare the results with and
without smearing. Our goal is to demonstrate that
smearing greatly speeds up topological optimization and
produces more robust optima. Though these optimization tasks fall in the realm of image analysis, we remind the reader that our framework is applicable anytime one wants to quickly optimize a topological functional on a simplicial complex. For instance, \textbf{STUMP} could be used to more rapidly, and perhaps robustly, compute a topological regularizer of a model representable as a simplicial complex. We have three synthetic
experiments:

\begin{enumerate}
	\item Double well: The image consists of two depressions, or wells, that have some overlap. The goal is to increase $H_0$ persistence and separate the wells. This is done by applying topological backpropagation to the critical vertices responsible for deaths in $H_0$, i.e. we create $H_0$ by raising a wall between the two wells, rather than making the wells lower.
	\item Sampled circle: The image consists of a sum of Gaussians centered at points sampled from a circle. The goal is to increase $H_1$ and fill in the circle. This is done by applying topological backpropagation to the critical vertices responsible for births in $H_1$, i.e. we want to create $H_1$ by making the circle appear earlier, rather than raising the center of the circle.
	\item Blobs: The image consists of some amorphous blobs connected by bridges at middling height. The goal is to decrease $H_0$, thereby better connecting the blobs. This is done by applying backpropagation to the critical vertices responsible for deaths in $H_0$, i.e. we want to decrease $H_0$ by deepening the bridges between them, rather than raising and flattening the blobs out.
\end{enumerate}

For our three experiments: (a) The persistence region of interest was $[-\infty,\infty,50,\infty]$ in birth-lifetime space, (b) The weighting $\alpha$ in the mixed-loss is $(1-1/P)$, where $P$ is the total number of pixels in the image. This balances the topological loss, whose gradient is supported on a relatively sparse set of pixels, and the MSE, whose gradient is supported on every pixel, (c) The learning rate is $5 \times 10^{-2}$, (d) We used the Adam optimizer \cite{kingma2014adam} with $10000$ steps, (e) Each pixel was perturbed independently by adding uniform noise in the range $[-\epsilon, \epsilon]$. The level of noise $\epsilon$ was $50$ for both the well and blobs experiment and $100$ for the circle experiment, (f)  For smeared loss, the $1$-Wasserstein norm was used to define the functional, although the $2$-Wasserstein norm also gives good results. For vanilla topological backpropagation, the $2$-Wasserstein norm was used, as the $1$-Wasserstein optima were very poor, and tended not to adjust the topology at all, (g) GUDHI \cite{gudhi:CubicalComplex} was used for all persistence computations. (h) Downsampling was done using method 3 described in Section \ref{sec:generatingapproximates}.

\begin{figure}[htb!]
	\begin{center}
		\includegraphics[scale=0.2]{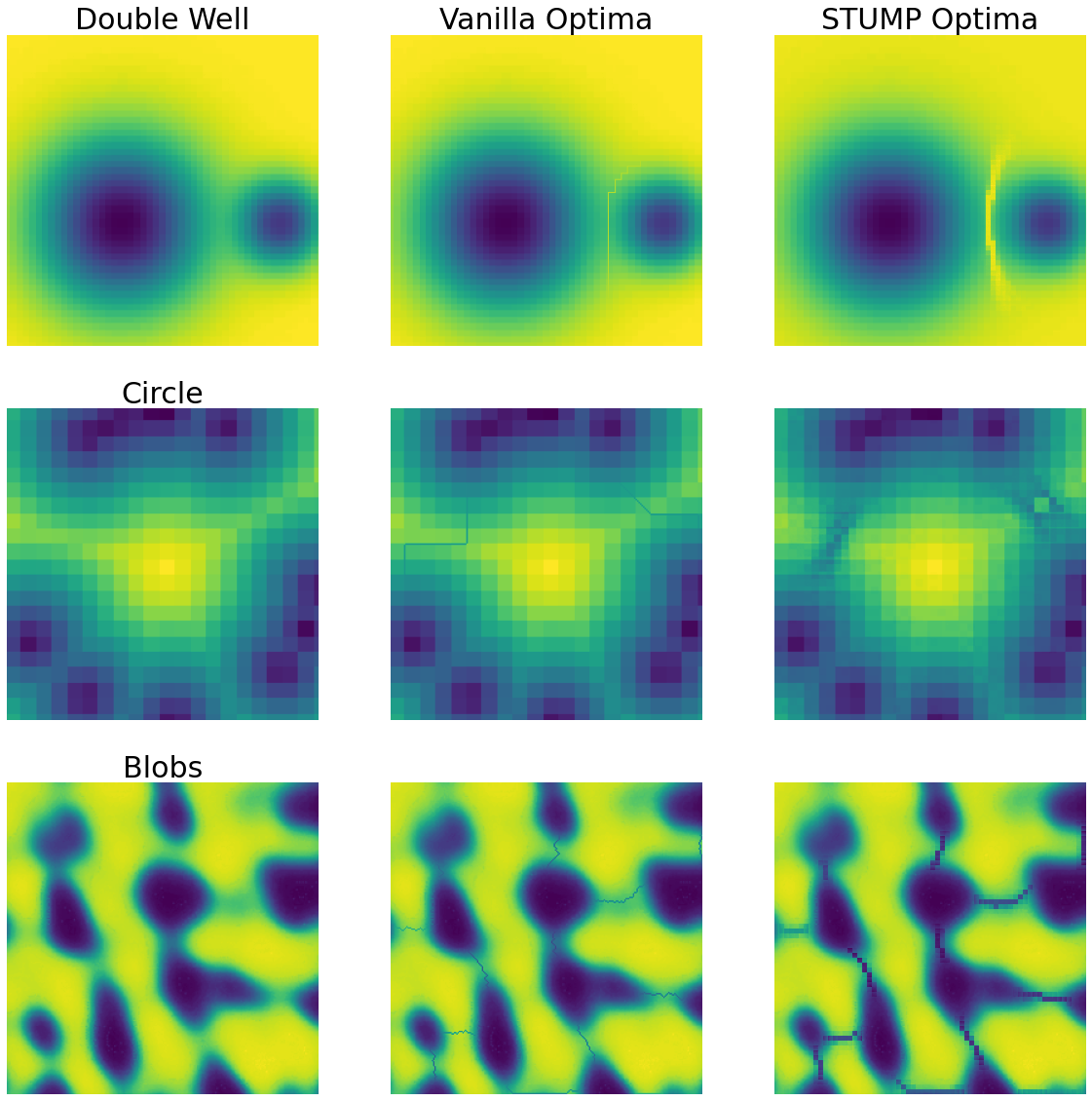}
	\end{center}
	\caption{A comparison of vanilla and \textbf{STUMP} optima for three optimization tasks. The images go from 0 (black) to 255 (yellow) in value.}
	\label{fig:opt_results}
\end{figure}

The results can be seen in Figure \ref{fig:opt_results}. We see in all three examples that the optima produced by \textbf{STUMP} look more stable and match closely with our intuition for what the goal of the optimization task should be. What remains is to compare the robustness and speed of vanilla and smeared topological backpropagation. 

\begin{figure}[htb!]
	\begin{center}
		\includegraphics[scale=0.4]{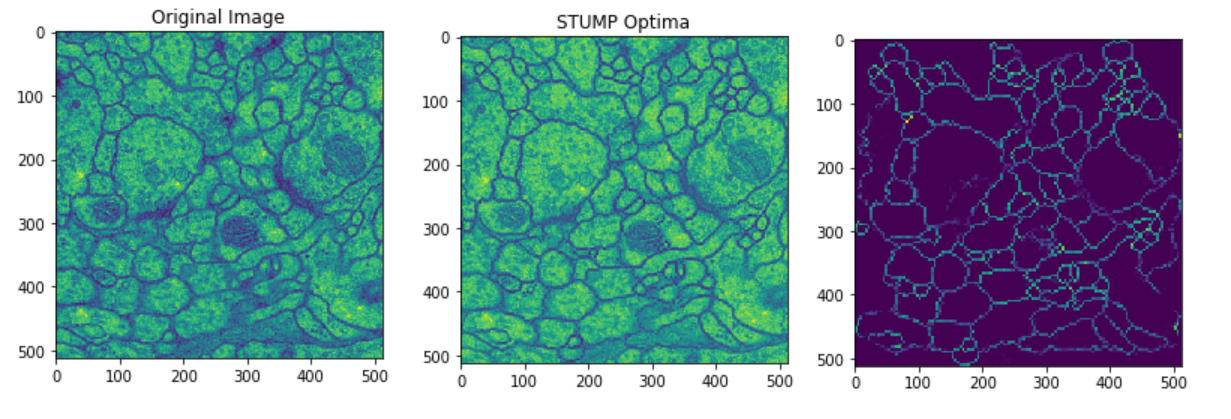}
	\end{center}
	\caption{\textbf{STUMP} applied to image from ISBI12 dataset. By neglecting local features, \textbf{STUMP} rapidly produces an altered image whose difference with the original reflects the segmentation.}
	\label{fig:cellsegmentation}
\end{figure}

We also have one non-synthetic experiment: cell segmentation. For this experiment, we only run the optimization with smearing, due to the unfeasible wait times of vanilla optimization. We consider the ISBI12 cell image data set taken from \cite{cardona2010integrated} that is also studied in \cite{hu2019topology}. In \cite{hu2019topology}, a topological component is added to the loss function of a neural network trained on supervised examples of image segmentations. Here our goal is unsupervised image segmentation. To make this a topological optimization task, we set the topological function to maximize $1$-dimensional homology with sufficiently large persistence (above $70$, for these examples), using a $4 \times 4$ downsampling filter, noise $\epsilon = 20$, by lowering the intensity of critical birth pixels. Intuitively, these one dimensional features are cell boundaries, and the effect of this optimization is to increase the pixel intensity along the boundaries. The advantage of using the downsampling filter, in addition to significant speedup, is that it erases the local $1$-dimensional features that do not correspond to cell boundaries. Figure \ref{fig:cellsegmentation} contains the original image, the \textbf{STUMP} optima, and the difference between them. Although run for only $5000$ descent steps, the difference image already contains the full topological segmentation. To address concerns that our methodology only works for mean-squared error, we performed this optimization for both MSE and binary cross-entropy. The results were substantially similar, and Figure \ref{fig:cellsegmentation} contains the binary cross-entropy optimum.

\subsection{Robustness}

\begin{figure}[htb!]
	\centering
	\includegraphics[width=1\linewidth]{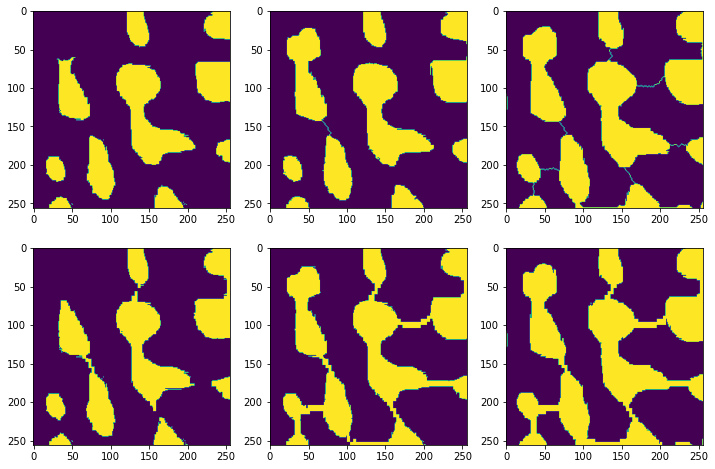}
	\caption{Results of the random walker segmentation algorithm on the vanilla (top row) and \textbf{STUMP} (bottom row) optima. The columns correspond to different choices of a thresholding hyperparameter.}
	\label{fig:randomwalker}
\end{figure}

A way of quantifying the extent to which components in the image are robustly connected is to feed the image to a segmentation algorithm and observe the connected components that it produces. We consider the random walker segmentation algorithm \cite{1704833}. Given an input image and a set of markers labelling phases, the algorithm marks an unknown pixel by considering a diffusion problem and labelling the unknown pixel with the label of the known marker whose probability of reaching the unknown pixel is highest. In our experiment, for a given $q \in [0, 1]$, we choose the markers to be the pixels whose value is in the top or bottom $q$ percent of the pixel values. Figure \ref{fig:randomwalker} shows the result of this procedure for the blobs optima from Figure \ref{fig:opt_results} and $q = 0.1, 0.2, 0.3$. We see that the \textbf{STUMP} optima in the bottom row is better able to keep the clusters connected across values of $q$.

\subsection{Speed}

\begin{figure}[htb!]
	\begin{center}
		\includegraphics[width=0.235\textwidth]{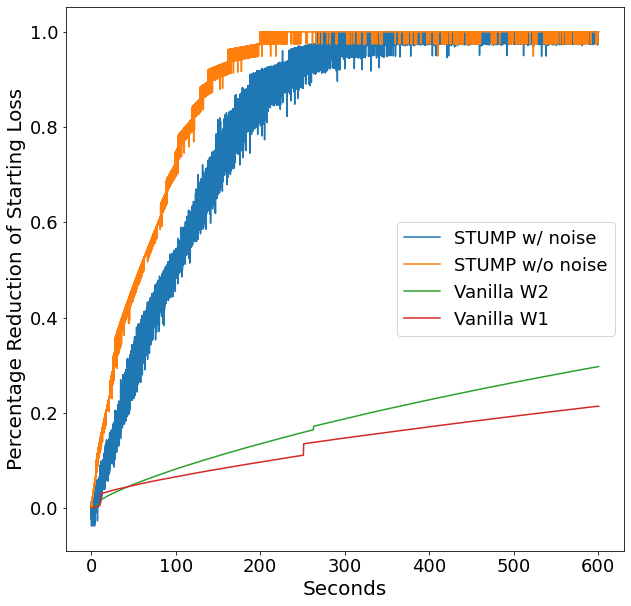}
		\includegraphics[width=0.235\textwidth]{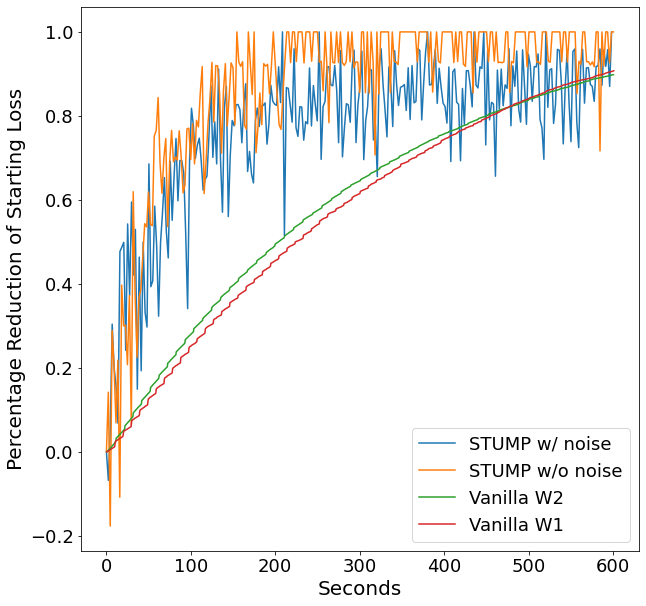}
	\end{center}
	\caption{Percentage of loss reduction as a function of time for the blobs image and uniform noise.}
	\label{fig:speedtestblobs}
\end{figure}

In all the preceding examples, downsampling was performed by considering adjacent $k \times k$ patches of the original image and, for each patch, applying a random element $\omega$ of $\Delta^{k^2-1}$, chosen uniformly. Hence, the downsampled image contained $k^{-2}$ as many pixels as the original image. In the associated optimization, this replaces the computation of $\nabla \Phi(\PD(f))$ with the faster computation of $\nabla \PD(f_\omega))$. Because of this, the vanilla wells, circle, and blobs experiments took $5015$, $2169$, and $3576$ seconds, respectively, while \textbf{STUMP} took $202$, $106$, and $195$ seconds.

Replacing the original gradient with a downsampled gradient certainly speeds up each step compared to vanilla optimization, but it remains to show that the loss function is reduced more rapidly. To this end, we now return to the third experiment regarding connecting blobs. Since the loss for this experiment consists of two non-negative terms, the mean squared error and the total persistence in a region, we may reasonably compare how quickly various types of optimization reduce the starting loss. In Figure \ref{fig:speedtestblobs}, we consider four types and plot the percentage of the original loss reduced by each optimization procedure as a function of time. In red and green, we show vanilla topological optimization where total persistence is measured using $W^1$ and $W^2$, respectively. We then consider the addition of stochastic downsampling in orange, where total persistence is measured using $W^1$. Finally, in blue, we add explicit noise before downsampling the image. The graph on the left corresponds to the blobs image shown in Figure \ref{fig:opt_results}. Within four minutes, both versions of our procedure have reduced the loss by about $90\%$ while the vanilla methods only manage to reduce around $25\%$ of the loss after $10$ minutes. 

One possible explanation for the dramatic increase in loss reduction in the blobs experiment is the large degree of homogeneity of this image. The second graph in Figure \ref{fig:speedtestblobs} corresponds to an identical optimization scheme for a different image. This new image was generated by sampling uniform noise between $0$ and $255$. For this experiment, we see a less extreme increase in loss reduction afforded by our procedure over vanilla optimization.

\section{Extensions}
\label{sec:extensions}
The methodology of smearing, and the \textbf{STUMP} pipeline, can also be applied to other settings and purposes.
\subsection{Critical Smears}
Strictly speaking, the method of topological optimization via smearing the loss function does not accomplish the task of topological backpropagation. That is, it works by considering gradients on many different persistence diagrams, as opposed to working with the gradient of the persistence diagram of the original function $f$. However, there is a way to use the ideas of smearing to this end as well, which we call \emph{critical smearing}.

In critical smearing, we compute the gradient of the original topological loss $\Phi(\PD(f))$, giving rise to a gradient on the persistence diagram $\PD(f)$. We then compute the persistence diagrams of many different functions of the form $(f+h)_{\omega}$, and \emph{transfer} the gradient from $\PD(f)$ to gradients on these approximate diagrams. We then pull back these transferred gradients to gradients on the \v{C}ech complex via persistence dot-critical vertex pairings, and finally back to gradients on $D$ via $\omega$, where the gradients are averaged to give a \emph{smeared gradient}. If the initial gradient on $\PD(f)$ is supported on a single dot, the resulting smeared gradient can be thought of as a fuzzy assignment of critical vertices for this dot. 

\begin{figure}[htb!]
	\begin{center}
		\includegraphics[scale=0.7]{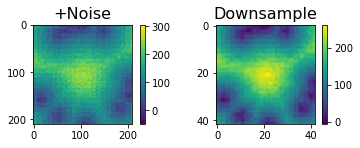}
	\end{center}
	\caption{Left: Circle with uniform noise added pixelwise. Right: Noisy image after pooling with $5 \times 5$ blocks.}
	\label{fig:circledwnsmpl}
\end{figure}

\begin{figure}[htb!]
	\begin{center}
		\includegraphics[scale=0.5]{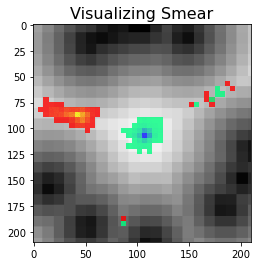}
	\end{center}
	\caption{Visualization of the critical smear corresponding to the underlying $1$-dimensional circular feature. The birth cells are in red, and the death cells are in blue.}
	\label{fig:circlesmear}
\end{figure}

There are many possible ways to define gradient transfer between persistence diagrams. We propose that this step be accomplished via finding a matching between the dots of two persistence diagrams, and having points in one diagram inherit the gradients of the points they are matched with. Fast matchings can be computed via the Sliced Wasserstein approximation of the Wasserstein distance (cf. \cite{carriere2017sliced}), and that is the approach we adopt here.
Consider again the circle in Figure \ref{fig:opt_results}, first column, second row. When we add uniform noise in $[-50,50]$ and subsequently downsample using $5 \times 5$ blocks, we obtain images as in Figure \ref{fig:circledwnsmpl}. If we set our loss function $\Phi$ to penalize dots in the persistence diagram with lifetime greater than $30$, add uniform noise in $[-50, 50]$, downsample via $5 \times 5$ blocks,  sample $1000$ times, and transfer gradients via Sliced Wasserstein (with $20$ projections), the critical smear can be seen in Figure \ref{fig:circlesmear}.

\subsection{Point Clouds}

It is relatively straightforward to adjust the above pipeline for topological backpropagation on point clouds. Downsampling can be accomplished by randomly sampling a subset of points, and error can be modeled by randomly perturbing the location of each point independently.

\section{Conclusion}
Our novel pipeline for topological optimization, \textbf{STUMP}, produces optima that are empirically more robust, and visually more intuitive, than the traditional method and with a considerably shorter computation time. The generalizability and parallelizability of gradient smearing opens the way to a host of promising interactions between applied topology and machine learning.
	
	\bibliographystyle{plain}
	\bibliography{topsimpbib}
\end{document}